\documentclass[twoside,11pt]{article}

\usepackage{blindtext}

\usepackage{dmlr2e}

\usepackage{textcomp}
\usepackage{xcolor}
\usepackage{booktabs}
\usepackage{multirow}
\usepackage{hhline}
\usepackage{hyperref}
\usepackage{textgreek}
\usepackage{amssymb}
\usepackage{bm,amsmath}
\usepackage{setspace}
\usepackage{adjustbox}
\usepackage{enumitem}
\usepackage{array} 
\usepackage{multirow} 
\usepackage{caption} 
\usepackage{subcaption} 

\usepackage{lastpage}
\dmlrheading{23}{\textit{The 12th International Conference on Learning Representations}, Vienna, Austria. DMLR@ICLR 2024.}{1-\pageref{LastPage}}{2/24; Revised 4/24}{4/24}{21-0000}{} 

\ShortHeadings{Data-centric Machine Learning Research - ICLR 2024}{}
\firstpageno{1}

\begin{document}

\begin{center}
    {\Large \textbf{PRE: Vision-Language Prompt Learning with Reparameterization Encoder}} \\[1.5cm]
\end{center}

\noindent 
\begin{tabular}{p{0.55\textwidth} p{0.45\textwidth}}
    \textbf{Thi Minh Anh Pham} & \texttt{m.pham@kingston.ac.uk} \\
    \small\textit{School of Computer Science and Mathematics} & \\
    \small\textit{Kingston University London} & \\[1em]
    
    \textbf{Duc-An Nguyen} & \texttt{an.nguyen@cs.ox.ac.uk} \\
    \small\textit{Department of Computer Science} & \\
    small\textit{University of Oxford} & \\[1em]
    
    \textbf{Cephas Svosve} & \texttt{cephas.svosve@maths.ox.ac.uk} \\
    \small\textit{Department of Mathematics} & \\
    \small\textit{University of Oxford} & \\[1em]
    
    \textbf{Vasileios Argyriou} & \texttt{vasileios.argyriou@kingston.ac.uk} \\
    \small\textit{School of Computer Science and Mathematics} & \\
    \small\textit{Kingston University London} & \\[1em]
    
    \textbf{Georgios Tzimiropoulos} & \texttt{g.tzimiropoulos@qmul.ac.uk} \\
    \small\textit{Department of Computer Science} & \\
    \small\textit{Queen Mary University of London} & \\
\end{tabular}

\vspace{1cm} 



\begin{abstract}
Large vision-language foundation models such as CLIP have demonstrated great potential in zero-shot transferability to downstream tasks. However, manual prompt engineering is the major challenge for deploying such models in practice since it requires domain expertise and extreme time. To avoid non-trivial prompt engineering, recent work Context Optimization (CoOp) introduced the concept of prompt learning to the vision domain using learnable textual tokens. While CoOp can achieve substantial improvements over manual prompts, its learned context is worse generalizable to wider unseen classes within the same dataset. In this work, we present Prompt Learning with Reparameterization Encoder (PRE) - a simple and efficient method that enhances the generalization ability of the learnable prompt to unseen classes in practical domains. Instead of directly optimizing the prompts, PRE employs a prompt encoder to reparameterize the input prompt embeddings, enhancing the exploration of domain-specific knowledge from few-shot data. Experiments and extensive ablation studies on 8 benchmarks demonstrate that our approach is an efficient method for prompt learning in vision-language foundation models. Specifically, PRE achieves a notable enhancement of 5.60\% in average accuracy on \textit{New} classes and 3\% in \textit{Harmonic mean} compared to CoOp in the 16-shot setting.\\
\textbf{Keywords}: prompt learning; domain specific data; few-shot learning; vision-language foundation models; CLIP. 
\end{abstract}

\section{Introduction}


In recent years, vision-language foundation models (VLMs) have brought new light on leveraging natural language supervision in visual recognition systems, enabling a wide exploration of open-set visual concepts \cite{li2021supervision, jia2021scaling, yao2021filip}. 
Notably, VLMs with contrastive learning, exemplified by Contrastive Language-Image Pretraining (CLIP) models \cite{radford2021learning}, have gained diverse visual concepts and rich cross-modal representations that hold great potential to be transferred to various tasks.
As these VLMs evolve, a pivotal question arises: How can the valuable knowledge from pretraining be effectively adapted to downstream tasks with domain-specific data?

\begin{table}[t]
\centering
\caption{Compared to existing methods on 8 datasets with 16-shot settings, PRE obtains a higher performance within good training time.}
\renewcommand{\arraystretch}{1.0}
\begin{adjustbox}{width=0.65\textwidth}
\begin{tabular}{lccccc}
\toprule
\multirow{2}{*}{\textbf{Methods}} & \multirow{2}{*}{\textbf{Prompts}} & \multicolumn{3}{c}{\textbf{Accuracy (\%)}} & \multirow{2}{*}{\textbf{Training Time}} \\
\cmidrule(lr){3-5} &  & \textbf{Base} & \textbf{New} & \textbf{H} &  \\
\midrule
CLIP & hand-crafted & 68.81 & 74.43 & 71.42 & - \\
\addlinespace[0.5ex]
\hline
\addlinespace[0.5ex]
CoOp & textual & 83.32 & 66.92 & 73.34 & 6ms/image \\
\addlinespace[0.25ex]
CoCoOp & textual+visual & 80.89 & 70.99 & 74.47 & 160ms/image \\
\addlinespace[0.25ex]
ProGrad & textual & 82.96 & 70.30 & 75.58 & 22ms/image \\
\addlinespace[0.25ex]
\midrule
{PRE} & textual & 82.14 & \textbf{71.90} & \textbf{76.22} & 6.2ms/image \\
\addlinespace[0.25ex]
\bottomrule
\end{tabular}
\end{adjustbox}
\end{table}

In the initial study \cite{radford2021learning}, prompt engineering has been utilized to add more meaningful context in textual class descriptions by using a set of manually selected prompts for the given task. For example, on the Oxford Pets dataset, employing a tuning prompt such as "a photo of a class, a type of pet." helps improve the accuracy performance \cite{yao2021cpt} \cite{jin2021good}. However, prompt engineering relies on trial and error, demanding significant human effort for word tuning - a slight change in wording could make a huge difference in performance, and does not guarantee the optimal prompts. 

Following research in NLP \cite{li2021prefix}, many recent works, beginning with Context Optimization(CoOp) \cite{zhou2022learning} introduced the concept of prompt learning to replace the manual prompts with a sequence of prompt tokens. Then, these prompt tokens are learned by minimizing the distance between the visual features and prompt-based text features using a few training examples to provide more flexibility in text encoding.
Despite significant improvements over CLIP, a noticeable problem with CoOp-based methods is the poor generalization to the unseen (\textit{New}) classes within the same practical domain.

The soft prompt optimization in CoOp attempts to learn separate prompt tokens solely through the pre-trained knowledge embedded in the fixed parameters of CLIP's text encoder. Because of the fixed over-parameterization of CLIP and lack of training examples, naive prompt tuning would lead to overfitting the seen (\textit{Base}) classes on specific datasets. 
Intuitively, we believe the values of prompt embeddings should be dependent on each other rather than independent, and their interdependence varies according to different domain-specific data. 
We need a trainable mechanism that jointly processes prompt embeddings. It could flexibly capture domain-relevant dependencies within prompt tokens beyond the constraints of the frozen text encoder. Incorporating these associate-learned prompt tokens into the text encoder would make the optimization easier to find a more contextually generalizable prompt specific to the particular dataset. 
Based on our hypothesis, we introduce Prompt Learning with a Reparameterization Encoder (PRE). Our main contributions are as follows:
\begin{itemize}[leftmargin=*]
\item Instead of directly learning the prompts, PRE reparameterizes the original prompt embeddings utilizing a prompt encoder before feeding them into the Text Encoder. Our prompt encoder incorporates a Bidirectional long short-term memory network (BiLSTM). This BiLSTM not only serves as a parameterizing network but also exploits the domain-specific long-range dependencies in the prompt sequence. We further adopt a residual connection for the prompt encoder to avoid forgetting the original knowledge encoded by the pre-trained CLIP. 
\item  We perform extensive ablation studies of PRE on eight classification datasets to analyze its characteristics. Specifically, several network architectures have been implemented in the prompt encoder, thus offering distinct benefits for different recognition tasks. The code is available at \href{https://github.com/minhanh151/PRE}{Github}.
\end{itemize}

\textbf{Main results:} We assess the performance of PRE through extensive experimentation on the base-to-new generalization setting across eight image classification datasets. The evaluation results in Table 1 highlight the efficiency and effectiveness of PRE. Our method demonstrated substantial accuracy improvement for the \textit{New} class compared to other methods while maintaining good training time.

\section{Related Work}
\subsection{Vision-language Foundation Models}\label{AA} 

The field of Vision-Language Models has experienced significant progress 
in forming robust representations that can be effectively transferred to various downstream tasks \cite{kamath2021mdetr, li2019simple, hong2020recurrent, kim2021vilt, bao2022vlmo}. 
A typical VLM comprises three components: an image encoder, a text encoder, and a loss function. 
Recent works such as CLIP \cite{radford2021learning}, ALIGN \cite{jia2021scaling}, and DeCLIP \cite{li2021supervision} bridge the vision-language modalities by learning text encoder and image encoder jointly with a contrastive loss, using large image-caption pairs datasets. Notably, CLIP showcases an impressive ability for zero-shot image recognition. Similar to the previous work CoOp and CoCoOp, we apply the pre-trained CLIP for knowledge transfer, aiming to facilitate the adaptation of such models in downstream datasets.

\subsection{Prompt Learning}\label{AA}
Prompt learning has emerged as a novel paradigm in NLP for exploiting pre-trained language models, gradually replacing the traditional fine-tuning transfer approach. The main idea of prompt learning is to formulate various NLP tasks as masked language modeling problems, adopting different templates (or prompts). 
In the context of Vision-Language Models like CLIP, human-crafted prompts based on class names are utilized to enable zero-shot visual recognition. Context Optimization (CoOp) \cite{zhou2022learning} extended soft prompt optimization to VLMs, where a set of prompts is learned and used as input to the text encoder alongside the class name. However, CoOp suffers from weak generalization, as the learned prompts tend not to capture domain-specific data well and perform poorly in novel classes. To address CoOp's generalization limitations, later work Conditional Context Optimization (CoCoOp) \cite{zhou2022conditional} proposes a dynamic approach. It employs a small neural network to produce a visual feature from each input image, which is then combined with the learned prompts, making them input-specific. However, CoCoOp's Meta-Net demands additional computation, which can be limiting, especially when dealing with large datasets or resource-constrained environments. 

Instead of directly learning the prompts like CoOp, PRE first passes the prompt tokens through a trainable encoder with a residual connection, enabling a flexible combination of the original prompt embeddings and embedding projections. This approach leads PRE to outperform CoOp, particularly in exploiting domain-specific data to handle unseen classes.


\section{Methodolgy}\label{AA}
\subsection{Preliminaries}\label{AA}
Our proposed method builds upon CLIP \cite{radford2021learning}, which is a well-known VLMl trained on an extensive dataset of 400 million image-text pairs. CLIP contains a visual encoder ($\phi$)  responsible for mapping images to visual embeddings and a textual encoder ($\theta$) used for embedding corresponding textual information.  During pretraining, CLIP LIP trains the image and text encoders by contrastive loss, which tries to maximize the similarity between matching pairs while minimizing the similarity with mismatching pairs.

\textbf{Prompt Engineering:} For downstream recognition tasks, CLIP performs the zero-shot inference by employing hand-engineered prompts to generate textual class embeddings. Given set $\mathcal{V}$ of $\mathcal{C}$ class names,  the class descriptions $\{\textbf{t}_c\}_{c=1}^C$ are generated with the manually designed prompt template, such as “$\texttt{a class of a \{class\_name\}}$”. Then the class descriptions are passed through the text encoder $\theta(\cdot)$ to compute the class-specific textual embeddings (weight): $\textbf{w}_i^{C}$ = $\theta(\textbf{t}_i^{C})$.
Given an image \textbf{x} along with its label $y$, the image features are extracted with the visual encoder $\phi(\cdot)$: \textbf{f} = $\phi(\textbf{x})$. 
\begin{equation}
    P(y=i\mid\textbf{x}) = \frac{\text{exp}(cos(\mathbf{w}_i,\mathbf{f})/\tau)}{\sum^{C}_{j=1}\text{exp}(cos(\mathbf{w}_j,\mathbf{f})/\tau)}
\end{equation}
where $cos(\cdot)$ denotes the cosine similarity and $\tau$ is a learnable temperature parameter in CLIP. Finally, the class label predicted for image \textbf{x} is given by $\Tilde{y} = arg_{max}~P(y|\textbf{x})$. 

\begin{figure*}[t]
    \center
    \includegraphics[width=0.925\textwidth]{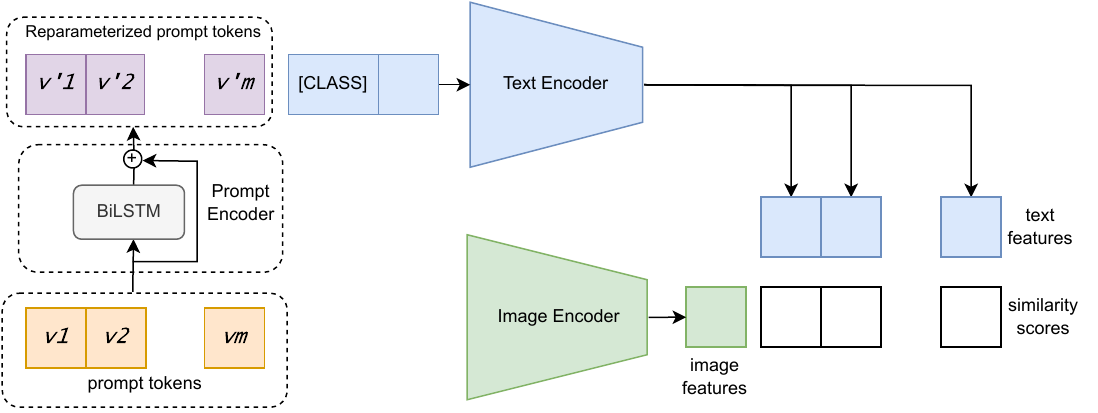}
    \caption{PRE is depicted in the following illustration: The original prompt embeddings $V$ undergo projection using a trainable prompt encoder $\mathcal{F}(\cdot)$ equipped with a residual connection. This process enables the modeling of domain-specific sequential dependencies within the input prompt embeddings and acquires a new mapping for the soft prompts.}
    \label{}
\end{figure*}


\textbf{Soft prompt learning:}
CLIP's reliance on human-crafted prompt templates for generating textual embeddings results in limited adaptability to downstream tasks. Recently, CoOp has utilized soft prompt learning approach from a few samples on the target task. Specifically, CoOp introduces $\textit{M}$ context vectors $V = \{\bm{v}_1, \bm{v}_2, ..., \bm{v}_M\}$ as the learnable prompt. The class embedding $\bm{c}_i$ of the $i$-th class is then concatenated with the learnable context vector $V$ for generating the prompts $\textbf{p}_i = \{\bm{v}_1, \bm{v}_2, ..., \bm{v}_M, \bm{c}_i\}$.
\begin{equation}
    P(y=i\mid\textbf{x}) = \frac{\text{exp}(cos(\theta(\mathbf{p}_i),\mathbf{f})/\tau)}{\sum^{C}_{j=1}\text{exp}(cos(\theta(\mathbf{p}_j),\mathbf{f})/\tau)}
\end{equation}
The prompts are learned by minimizing the cross-entropy loss:
\begin{equation}
    \mathcal{L}_{VL} = - \sum^{C}_{c=1} \text{log}~P(c|\mathbf{x})y_c
\end{equation}

\subsection{Prompt Learning with Reparameterization Encoder}\label{AA}
While excelling in \textit{Base} classes (68.81\% CLIP vs 83.32\% CoOp), soft prompt learning exhibits suboptimal results in novel classes (74.43\% CLIP vs 66.92\% CoOp), Table 1. 

 In this work, we propose a more adaptable parameterization of soft prompts, achieved through the utilization of a prompt encoder (Fig. 1). This encoder can be trained on the downstream task to enable domain-specific modifications to the prompt embeddings before forwarding them into the fixed text encoder. Specifically, we project the original prompt embeddings $V$ consisting of $\mathbf{}{M}$ prompt tokens $\{\bm{v}_1, \bm{v}_2, ..., \bm{v}_M\}$ through the prompt encoder into a reparameterized sequence $\bm{\Tilde{V}}$ as follows:
 \begin{equation}
    \bm{\Tilde{V}} = [\bm{\Tilde{v}}_1, \bm{\Tilde{v}}_2, ..., \bm{\Tilde{v}}_M] = [\mathcal{F}({\bm{v}_1}), \mathcal{F}({\bm{v}_2}), ..., \mathcal{F}({\bm{v}_M})]
\end{equation}
where $\mathcal{F}(\cdot)$ represents the reparameterization function of the prompt encoder, which consists of a network $\varphi(\cdot)$ with a residual connection. $\mathcal{F}(\cdot)$ is applied to each prompt token:
\begin{equation}
    \mathcal{F}({\bm{v}_i}) = \varphi({\bm{v}_i}) + \bm{v}_i, i \in \{1...M \} 
\end{equation}

\begin{figure}[t]
    \center
    \includegraphics[width=0.5\textwidth]{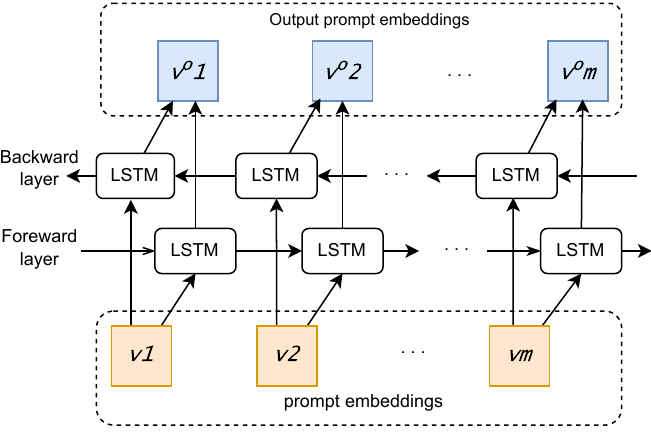}
    \caption{Illustration of one-layer Bidirectional LSTM architecture used in Prompt Encoder.}
    \label{}
\end{figure}

Network $\varphi(\cdot)$ acts as an adaptable mechanism that establishes associations between prompt embeddings and enables task-relevant reparameterization in the prompt embeddings. Different encoder network architectures exhibit varying effects on the model's adaptability (Ablation Studies). In PRE, we propose a one-layer BiLSTM architecture, as in Fig. 2. BiLSTM processes the input prompt sequence in both forward and backward directions simultaneously. By considering both backward and forward context, it enhances the modeling of the domain-specific sequential dependencies in the projected prompt tokens. The encoder itself has a skip connection. It enables the model to be more flexible in combining the original embedding of each prompt token with the mapping representation derived from the network $\varphi(\cdot)$ (Equation (5)). By employing this residual-style blending, PRE can flexibly combine the original knowledge encoded from CLIP and the newly learned knowledge acquired from the few-shot training examples through the BiLSTM network.

Once the projected context sequence $\bm{\Tilde{V}}$ is obtained from the prompt encoder, it is merged with the class token embedding $\bm{c}_i$. This combined input is then fed into the text encoder $\theta(\cdot)$, resulting in the generation of a class-specific encoded adaptable prompt $\textbf{t}^r_i = \theta(\textbf{r}_i)$.
\begin{equation}
    P(y=i\mid\textbf{x}) = \frac{\text{exp}(cos(\mathbf{t}^r_i,\mathbf{f})/\tau)}{\sum^{C}_{j=1}\text{exp}(cos(\mathbf{t}^r_j,\mathbf{f})/\tau)}
\end{equation}
\textbf{Training}: We train the prompt embeddings $V$ and the encoder parameters $\mathcal{F}(\cdot)$ on the downstream task, while preserving all other parameters fixed. The training objective is to maximize the log-likelihood of correct output y given the encoded learnable prompt $\textbf{t}^r_c$:
\begin{equation}
    \mathcal{L} = - \sum^{C}_{c=1} \text{log}~p(c|\mathbf{x};\mathbf{t}^r_c)y_c
\end{equation}
The gradients can be back-propagated through the
text encoder $\theta(\cdot)$ and the prompt encoder $\mathcal{F}(\cdot)$. This differential optimization not only utilizes the pre-trained knowledge stored in the fixed text encoder's parameters but also takes advantage of the prompt embeddings projected by the reparameterizing encoder. This combination guides the gradient towards convergence on more generalizable prompts to downstream tasks.

\section{Experiments}
Following CLIP, CoOp, and CoCoOp, we mainly evaluate the accuracy of our proposed method based on generalization from base-to-new classes within a dataset.

\textbf{Datasets:} The base-to-new generalization is conducted on 8 datasets, namely: Caltech101 for generic object classification; Oxford-Pets, Stanford Cars, Flowers102, Food101, FGVC Aircraft for fine-grained visual categorization; DTD for texture classification; and EuroSAT for satellite image classification.

\textbf{Models:} Our implementation is based on
CoOp’s code \cite{zhou2022learning} with the CLIP model. The experiments were conducted based on the vision backbone ViT-B/16 image encoder \cite{dosovitskiy2020image}. Similar to CoCoOp, we set the prompt tokens to 4 and initialize the context vectors using the template "a photo of a []". The class names are inserted at the end of these random templates. The final performance is averaged over three random seeds. 

\begin{table*}[t]
\centering
\caption{Comparison with existing methods in base-to-new generalization setting with ViT-B/16 as the backbone. The context length M is 4 with the 16-shot samples from the base classes. H: Harmonic mean (to highlight the generalization trade-off \cite{xian2017zero}).}

\begin{subtable}{0.33\textwidth}
\centering
\caption{Average over 8 datasets.}
\begin{adjustbox}{width=0.9\textwidth}
\begin{tabular}{@{}lcc|c@{}}
\toprule
& Base & New & H \\
\midrule
CLIP & 68.81 & \textbf{74.43} & 71.42  \\
\addlinespace[0.25ex]
CoOp & \textbf{83.32} & 66.92 & 73.34 \\
\addlinespace[0.25ex]
CoCoOp & 80.89 & 70.99 & 74.47 \\
\addlinespace[0.25ex]
ProGrad & 82.96 & 70.30 & 75.58 \\
\addlinespace[0.25ex]
\midrule
PRE & 82.02 & 71.90 & \textbf{76.22} \\
\bottomrule
\addlinespace[2ex]
\end{tabular}
\end{adjustbox}
\end{subtable}%
\begin{subtable}{0.33\textwidth}
\centering
\caption{Caltech101.}
\begin{adjustbox}{width=0.9\textwidth}
\begin{tabular}{@{}lcc|c@{}}
\toprule
& Base & New & H \\
\midrule
CLIP & 96.84 & \textbf{94.00} & 95.40 \\
\addlinespace[0.25ex]
CoOp & \textbf{98.11} & 93.02 & 95.50 \\
\addlinespace[0.25ex]
CoCoOp & 97.96 & 93.81 & 95.84 \\
\addlinespace[0.25ex]
ProGrad & 98.02 & 93.89 & \textbf{95.91} \\
\addlinespace[0.25ex]
\midrule
PRE & 98.00 & 93.50 & 95.70 \\
\bottomrule
\addlinespace[2ex]
\end{tabular}
\end{adjustbox}
\end{subtable}%
\begin{subtable}{0.33\textwidth}
\centering
\caption{OxfordPets.}
\begin{adjustbox}{width=0.9\textwidth}
\begin{tabular}{@{}lcc|c@{}}
\toprule
& Base & New & H \\
\midrule
CLIP & 91.17 & 97.26 & 94.12 \\
\addlinespace[0.25ex]
CoOp & 94.24 & 96.66 & 95.43 \\
\addlinespace[0.25ex]
CoCoOp & 95.20 & \textbf{97.69} & 96.43 \\
\addlinespace[0.25ex]
ProGrad & 95.07 & 97.63 & 96.33 \\
\addlinespace[0.25ex]
\midrule
PRE & \textbf{95.27} & 97.61 & \textbf{96.43} \\
\bottomrule
\addlinespace[2ex]
\end{tabular}
\end{adjustbox}
\end{subtable}

\begin{subtable}{0.33\textwidth}
\centering
\caption{StanfordCars.}
\begin{adjustbox}{width=0.9\textwidth}
\begin{tabular}{@{}lcc|c@{}}
\toprule
& Base & New & H \\
\midrule
CLIP & 63.37 & \textbf{74.89} & 68.85 \\
\addlinespace[0.25ex]
CoOp & \textbf{76.20} & 67.14 & 71.38 \\
\addlinespace[0.25ex]
CoCoOp & 70.49 & 73.59 & 72.01 \\
\addlinespace[0.25ex]
ProGrad & 76.68 & 68.63 & 72.43 \\
\addlinespace[0.25ex]
\midrule
PRE & 75.83 & 69.90 & \textbf{72.74} \\
\bottomrule
\addlinespace[2ex]
\end{tabular}
\end{adjustbox}
\end{subtable}%
\begin{subtable}{0.33\textwidth}
\centering
\caption{Flowers102.}
\begin{adjustbox}{width=0.9\textwidth}
\begin{tabular}{@{}lcc|c@{}}
\toprule
& Base & New & H \\
\midrule
CLIP & 72.08 & \textbf{77.80} & 74.83 \\
\addlinespace[0.25ex]
CoOp & \textbf{97.63} & 66.55 & 79.15 \\
\addlinespace[0.25ex]
CoCoOp & 94.87 & 71.75 & 81.71 \\
\addlinespace[0.25ex]
ProGrad & 95.54 & 71.87 & \textbf{82.03} \\
\addlinespace[0.25ex]
\midrule
PRE & 96.04 & 70.80 & 81.64 \\
\bottomrule
\addlinespace[2ex]
\end{tabular}
\end{adjustbox}
\end{subtable}%
\begin{subtable}{0.33\textwidth}
\centering
\caption{Food101.}
\begin{adjustbox}{width=0.9\textwidth}
\begin{tabular}{@{}lcc|c@{}}
\toprule
& Base & New & H \\
\midrule
CLIP & 90.10 & 91.22 & 90.66 \\
\addlinespace[0.25ex]
CoOp & 89.44 & 86.50 & 87.95 \\
\addlinespace[0.25ex]
CoCoOp & 90.70 & 91.29 & 90.99 \\
\addlinespace[0.25ex]
ProGrad & 90.37 & 89.59 & 89.98 \\
\addlinespace[0.25ex]
\midrule
PRE & \textbf{90.96} & \textbf{91.46} & \textbf{91.21} \\
\bottomrule
\addlinespace[1.75ex]
\end{tabular}
\end{adjustbox}
\end{subtable}%

\begin{subtable}{0.33\textwidth}
\centering
\caption{FGVCAircraft.}
\begin{adjustbox}{width=0.9\textwidth}
\begin{tabular}{@{}lcc|c@{}}
\toprule
& Base & New & H \\
\midrule
CLIP & 27.19 & \textbf{36.29} & 31.09 \\
\addlinespace[0.25ex]
CoOp & 39.24 & 23.49 & 29.39 \\
\addlinespace[0.25ex]
CoCoOp & 33.41 & 23.71 & 27.74 \\
\addlinespace[0.25ex]
ProGrad & \textbf{40.54} & 27.57 & 32.82 \\
\addlinespace[0.25ex]
\midrule
PRE & 35.63 & 32.43 & \textbf{34.53} \\
\bottomrule
\addlinespace[1.5ex]
\end{tabular}
\end{adjustbox}
\end{subtable}%
\begin{subtable}{0.33\textwidth}
\centering
\caption{DTD.}
\begin{adjustbox}{width=0.9\textwidth}
\begin{tabular}{@{}lcc|c@{}}
\toprule
& Base & New & H \\
\midrule
CLIP & 53.24 & \textbf{59.90} & 56.37 \\
\addlinespace[0.25ex]
CoOp & \textbf{80.17} & 47.54 & 59.69 \\
\addlinespace[0.25ex]
CoCoOp & 77.01 & 56.00 & \textbf{64.85} \\
\addlinespace[0.25ex]
ProGrad & 77.35 & 52.35 & 62.44 \\
\addlinespace[0.25ex]
\midrule
PRE & 77.84 & 53.93 & 63.70 \\
\bottomrule
\addlinespace[1.5ex]
\end{tabular}
\end{adjustbox}
\end{subtable}%
\begin{subtable}{0.33\textwidth}
\centering
\caption{EuroSAT.}
\begin{adjustbox}{width=0.9\textwidth}
\begin{tabular}{@{}lcc|c@{}}
\toprule
& Base & New & H \\
\midrule
CLIP & 56.48 & 64.05 & 60.03 \\
\addlinespace[0.25ex]
CoOp & \textbf{91.54} & 54.44 & 68.28 \\
\addlinespace[0.25ex]
CoCoOp & 87.49 & 60.04 & 71.21 \\
\addlinespace[0.25ex]
ProGrad & 90.11 & 60.89 & 72.67 \\
\addlinespace[0.25ex]
\midrule
PRE & 86.23 & \textbf{64.47} & \textbf{73.78} \\
\bottomrule
\addlinespace[1.5ex]
\end{tabular}
\end{adjustbox}
\end{subtable}%

\label{tab:comparison}
\end{table*}

\textbf{Training Details:} We maintain consistency with CoOp \cite{zhou2022learning} and CoCoOp \cite{zhou2022conditional} in terms of training epochs and training procedures which adopted the SGD optimizer with 0.002 initial learning rate, CosineAnnealingLR schedule. 
We conducted all training and testing on two NVIDIA GeForce RTX 3080 Ti GPUs.

\textbf{Baselines:} We present the results of the PRE method to compare its performance with CLIP (hand-crafted prompts) and three soft prompt learning methods including CoOp, CoCoOp, and ProGrad.


\subsection{Generalization From Base-to-New Classes}\label{AA}
On each of the 8 datasets, we split the classes equally into two groups, one as \textit{Base} classes and the other as \textit{New} classes. Learning-based models, i.e., CoOp, CoCoOp and PRE, are trained using only the \textit{Base} classes while evaluation is conducted on the \textit{Base} and \textit{New}
classes separately to assess the generalization capabilities.
The detailed results on 16-shot settings, $M$ = 4 learnable prompts on ViT-B/16 are shown in Table 2.

\begin{table*}[b]
\centering
\caption{Comparison in the base-to-new setting with different K-shot samples in terms of the average performance among all 8 datasets with backbones ViT-B/16.}
\begin{adjustbox}{width=1\textwidth}
\begin{tabular}{lcccc|ccc|ccc}
\toprule
\multirow{2}{*}{Methods} & \multirow{2}{*}{Prompts} & \multicolumn{3}{c}{$K$ = 4} & \multicolumn{3}{c}{$K$ = 8} & \multicolumn{3}{c}{$K$ = 16} \\
\cmidrule(lr){3-11} & & Base & New & H & Base & New & H  & Base & New & H \\
\midrule
CoOp & textual & \textbf{79.33} & 66.02 & 71.47 & \textbf{81.67} & 66.29 & 71.89 & \textbf{83.32} & 66.92 & 73.34 \\ 
\addlinespace[0.25ex]
CoCoOp & textual+visual & 76.51 & \textbf{71.48} & 73.68 & 78.67 & 70.78 & 74.14 & 80.89 & 70.99 & 74.47 \\
\addlinespace[0.25ex]
ProGrad & textual & 79.15 & 70.40 & 74.11 & 80.55 & 70.84 & 75.00 & 80.96 & 70.30 & 75.58 \\
\addlinespace[0.25ex]
\midrule
{PRE} & textual & 78.62 & 71.31 & \textbf{74.46} & 79.84 & \textbf{71.49} & \textbf{75.07} & 82.02 & \textbf{71.90} & \textbf{76.22} \\
\bottomrule
\end{tabular}
\end{adjustbox}
\end{table*}

\textbf{PRE Significantly Narrows Generalization Gap.}
 Compared with existing methods, our proposed PRE method achieves the highest \textit{New} performance on three out of eight datasets. In direct comparison with CoOp, PRE demonstrates significant improvements in accuracy for unseen classes. Specifically, the accuracy increases more than 5\% from 67.14\% to 71.90\%, substantially narrowing the gap between soft prompt learning with manual prompts. The results confirm that our reparameterizing encoder $\mathcal{F}(\cdot)$ coupled with a residual connection, proficiently captures task-relevant dependencies within the initial prompt embeddings. This associate-learned projected prompt sequence navigates a more efficient optimization process to find a contextually generalizable prompt specific to the downstream domain. Furthermore, when compared to both CoCoOp and ProGrad, PRE showcases a relatively improved performance in novel classes - averaging 71.90\% against CoCoOp's 70.99\% and ProGrad's 70.30\%. This consistent enhancement in novel classes' performance across various classes demonstrates PRE's remarkable ability to balance performance across a wide spectrum of domains.

\textbf{Regarding the Harmonic mean (represents the generalization trade-off \cite{xian2017zero}), PRE outperforms all other methods.} 
Our proposed method consistently achieves the highest Harmonic mean across five out of eight datasets. On average, it surpasses CLIP by 5\%, CoOp by 3\%, CoCoOp by nearly 2\%, and ProGrad by 1\%. Besides the performance improvements in novel classes, this is partly attributed to the fact that PRE maintains a good \textit{Base} classes' performance. Specifically, PRE achieves a remarkable 6.7\% higher Base accuracy than CLIP and surpasses CoCoOp on six out of eight datasets with an average accuracy of 82.02\% vs 80.89\%. This performance improvement in seen classes comes from the efficacy of the residual connection, which dynamically balances and blends knowledge from pretrained CLIP and the newly acquired insights from the prompt encoder based on few-shot samples. 

\textbf{Various K-shot samples}:
Table 3 summarizes the average performance across all 8 datasets, considering various K-shot samples on ViT-B/16. Similar to the observations in the 16-shot settings, PRE consistently achieves a higher average Harmonic mean than existing methods. CoOp still achieves the best performance regarding \textit{Base} classes while obtaining the worst \textit{New} class performance in all K-shot samples. The performance gap between PRE and others is most significant when K = 16 shots and tends to diminish as K decreases. This phenomenon is because the reparameterizing encoder $\mathcal{F}(\cdot)$ in PRE requires a certain number of data samples to fully explore its potential in learning the optimal prompt embeddings.


\subsection{Ablation Studies}\label{AA}
In this subsection, we conducted ablation studies to investigate the effectiveness of different components in PRE. 

\textbf{Parameter-efficiency of PRE:} The total number of trainable parameters in PRE consists of 1) trainable prompt embeddings, and 2) reparameterizing encoder network. Our encoder network $\mathcal{F}(\cdot)$ contains a one-layer Bidirectional LSTM. Let $M$ be the number of prompt tokens and $d$ be the dimensionality of model embeddings, the BiLSTM network has an input size of $d$ and a hidden size of $d/2$, and it incorporates two separate LSTM units for the forward and backward direction. 
For each LSTM unit, the weight matrix for input-to-hidden connections has a shape of $4 \times d \times d/2$, and the weight matrix for hidden-to-hidden connections has a shape of $4 \times d/2 \times d/2$. Taking into account both the forward and backward LSTM units, we have $3 \times d \times d$ parameters of the reparameterizing encoder network and $d \times M$ soft prompt parameters. 
As the number of prompt tokens $M$ varies from 1 to 16, the number of trainable parameters in PRE remains relatively low, resulting in efficient training time compared to other methods (Table 1).

\begin{figure}[t]
    \centering
    \includegraphics[width=0.57\textwidth]{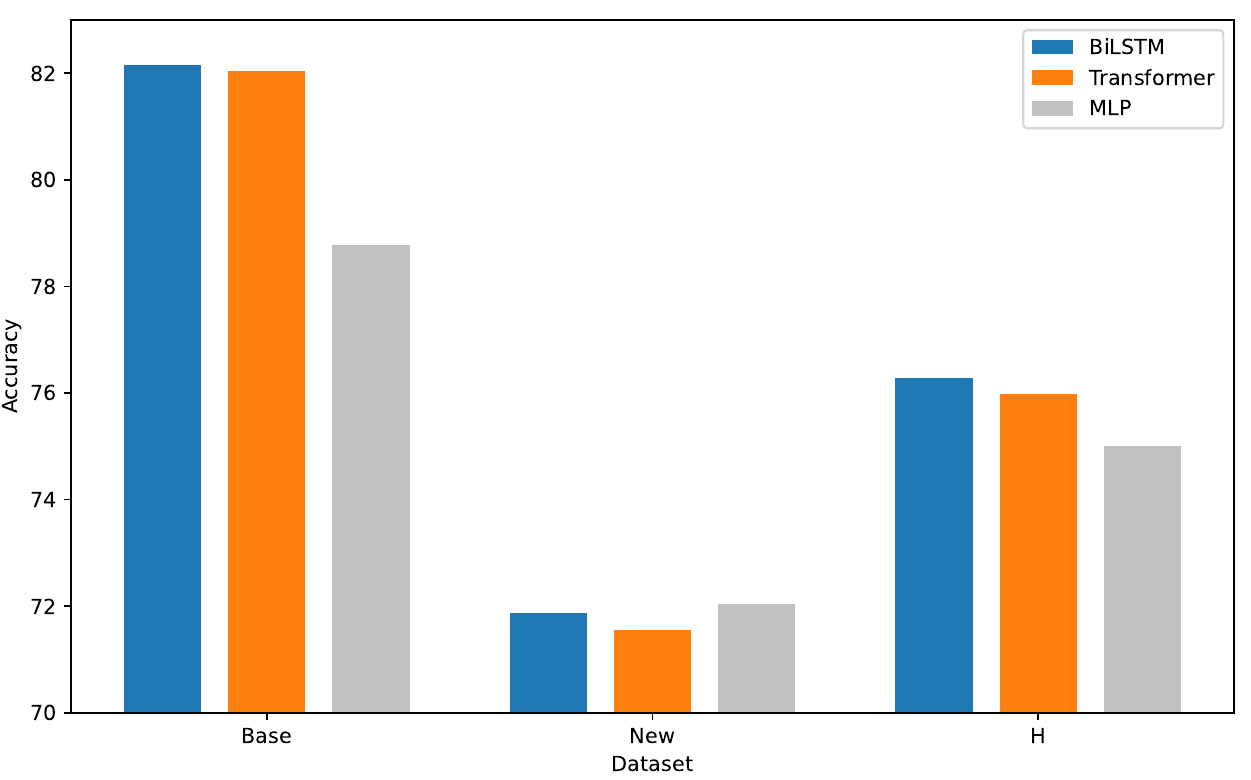}
    \caption{The impact of different reparameterizing encoder network architectures on average Accuracy on \textit{Base}, \textit{New} and \textit{H} over 8 datasets on the 16-shot setting.}
    \label{fig:my_label}
\end{figure}

\textbf{Effects of different reparameterization encoder network architectures:} To gain a deeper understanding of the reparameterizing mechanism and how its choice of network architecture impacts the model's performance, we have implemented a range of network architectures in the prompt encoder including a bottle-neck MLP, Transformer Encoder network, and BiLSTM, all equipped with a residual connection. All three networks lead to improvements in novel classes and the Harmonic mean compared to the CoOp method (Table 4). These results underscore the efficacy of the prompt reparameterization technique for adapting VLMs, although the outcomes vary based on the chosen network architecture. We observe that the Transformer Encoder network follows a similar trend to the BiLSTM, albeit with slightly lower performance. While MLP-based reparameterization demonstrates an improvement in \textit{New} class accuracy,  it comes with a significant decline in performance for the \textit{Base} class. This is due to bottleneck MLPs' inherent limitations in capturing long-range sequential dependencies and contextual information within prompt embeddings. As feedforward networks, MLPs process tokens independently, neglecting their relationships. This lack of contextual understanding may result in less contextually rich prompt embeddings. Among three architectures, BiLSTM has the most promising results and isn't affected by specific hyperparameter choices, making them easier to train and less prone to overfitting in few-shot training examples.

\begin{table}[t]
\centering
\caption{Comparison in different encoder network architectures with vs. without Residual connection in terms of the average \textit{Base}, \textit{New}, and \textit{H} performance in the 16-shot setting.}
\renewcommand{\arraystretch}{1.0}
\begin{adjustbox}{width=0.5\textwidth}
\begin{tabular}{ccccc}
\toprule
\multirow{2}{*}{Encoder Network} & \multirow{2}{*}{\shortstack{Residual\\connection}} & \multicolumn{3}{c}{Accuracy (\%)} \\
\cmidrule(lr){3-5} & & Base & New & H \\
\midrule
BiLSTM & \textbf{Yes} &82.02 & 71.90 & 76.22 \\
\addlinespace[0.25ex]
BiLSTM & No & 80.51 & 70.57 & 74.67 \\
\addlinespace[0.25ex]
 \cmidrule(lr){3-5}&  & -1.64 & -1.31 & -1.61 \\
\addlinespace[0.25ex]
\midrule
{Transformer Encoders} & \textbf{Yes} & 82.05 & 71.55 & 75.98 \\
\addlinespace[0.25ex]
{Transformer Encoders} & No & 82.01 & 69.34 & 74.54 \\
\addlinespace[0.25ex]
 \cmidrule(lr){3-5}&  & -0.04 & -2.21 & -1.44 \\
 \midrule
{Bottleneck MLP} & \textbf{Yes} & 78.77 & 72.04 & 75.00 \\
\addlinespace[0.25ex]
{Bottleneck MLP} & No & 74.54 & 68.88 & 71.16 \\
\addlinespace[0.25ex]
 \cmidrule(lr){3-5}&  & -4.23 & -3.16 & -3.84 \\
\bottomrule
\end{tabular}
\end{adjustbox}
\end{table}

\textbf{Residual connection is an important component in PRE:} 
According to Table 4, the performance of both the \textit{Base}, \textit{New}, and \textit{H} decreases when the skip connection is removed from the prompt encoder in all network architectures. Notably, MLP exhibits a significant drop of up to 4.23\% in accuracy for \textit{Base} classes and 3.16\% in the \textit{New} class performance, demonstrating a significant vulnerability in handling input prompt embeddings. This is because the MLPs are inherently shallow networks without inherent memory mechanisms to capture sequential dependencies. As a result, when the residual connection is removed, without the complementary from the original prompt embeddings, the MLP lacks the ability to effectively propagate contextual information through the layers, leading to a more significant performance degradation.

\textbf{Limitations and Future Works:} 
In terms of \textit{New} class performance, PRE's performance lags behind that of CLIP in 5 out of the 8 datasets (as seen in Table 2).
In terms of future work, one direction is to further develop the reparameterizing prompt encoder network with potentially a more efficient architecture that can enhance the model's generalizability. 
Furthermore, a promising avenue is to investigate the influence of our reparameterizing encoder within the context of other VLMs tuning methods such as adapter-based. We believe that with targeted modifications, our approach could potentially be integrated to accelerate certain aspects of these models, and this will be explored in future research.

\section{Conclusions}
In this paper, we presented PRE - a practical soft prompt learning method for Vision Language adaptation with limited data and resource constraints. Our method demonstrates its efficiency in achieving improved generalization performance over prior works while maintaining performance on the seen classes. We hope that our approach could be explored in collaboration with other well-established methods in the future, thereby contributing to the overall enhancement of adaptability within the realm of Vision Language models.

\section*{Acknowledgement}
This project has received funding from the European Union’s Horizon Europe research and innovation programme under grant agreement No. 101135800 (RAIDO).


\begingroup
\setstretch{1.22}
\bibliography{c_refercence}

\begin{thebibliography}{18}
\providecommand{\natexlab}[1]{#1}
\providecommand{\url}[1]{\texttt{#1}}
\expandafter\ifx\csname urlstyle\endcsname\relax
  \providecommand{\doi}[1]{doi: #1}\else
  \providecommand{\doi}{doi: \begingroup \urlstyle{rm}\Url}\fi

\bibitem[Bao et~al.(2022)Bao, Wang, Dong, Liu, Mohammed, Aggarwal, Som, Piao, and Wei]{bao2022vlmo}
Hangbo Bao, Wenhui Wang, Li~Dong, Qiang Liu, Owais~Khan Mohammed, Kriti Aggarwal, Subhojit Som, Songhao Piao, and Furu Wei.
\newblock Vlmo: Unified vision-language pre-training with mixture-of-modality-experts.
\newblock \emph{Advances in Neural Information Processing Systems}, 35:\penalty0 32897--32912, 2022.

\bibitem[Dosovitskiy et~al.(2020)Dosovitskiy, Beyer, Kolesnikov, Weissenborn, Zhai, Unterthiner, Dehghani, Minderer, Heigold, Gelly, et~al.]{dosovitskiy2020image}
Alexey Dosovitskiy, Lucas Beyer, Alexander Kolesnikov, Dirk Weissenborn, Xiaohua Zhai, Thomas Unterthiner, Mostafa Dehghani, Matthias Minderer, Georg Heigold, Sylvain Gelly, et~al.
\newblock An image is worth 16x16 words: Transformers for image recognition at scale.
\newblock \emph{arXiv preprint arXiv:2010.11929}, 2020.

\bibitem[Hong et~al.(2020)Hong, Wu, Qi, Rodriguez-Opazo, and Gould]{hong2020recurrent}
Yicong Hong, Qi~Wu, Yuankai Qi, Cristian Rodriguez-Opazo, and Stephen Gould.
\newblock A recurrent vision-and-language bert for navigation.
\newblock \emph{arXiv preprint arXiv:2011.13922}, 2020.

\bibitem[Jia et~al.(2021)Jia, Yang, Xia, Chen, Parekh, Pham, Le, Sung, Li, and Duerig]{jia2021scaling}
Chao Jia, Yinfei Yang, Ye~Xia, Yi-Ting Chen, Zarana Parekh, Hieu Pham, Quoc Le, Yun-Hsuan Sung, Zhen Li, and Tom Duerig.
\newblock Scaling up visual and vision-language representation learning with noisy text supervision.
\newblock In \emph{International conference on machine learning}, pages 4904--4916. PMLR, 2021.

\bibitem[Jin et~al.(2021)Jin, Cheng, Shen, Chen, and Ren]{jin2021good}
Woojeong Jin, Yu~Cheng, Yelong Shen, Weizhu Chen, and Xiang Ren.
\newblock A good prompt is worth millions of parameters: Low-resource prompt-based learning for vision-language models.
\newblock \emph{arXiv preprint arXiv:2110.08484}, 2021.

\bibitem[Kamath et~al.(2021)Kamath, Singh, LeCun, Synnaeve, Misra, and Carion]{kamath2021mdetr}
Aishwarya Kamath, Mannat Singh, Yann LeCun, Gabriel Synnaeve, Ishan Misra, and Nicolas Carion.
\newblock Mdetr-modulated detection for end-to-end multi-modal understanding.
\newblock In \emph{Proceedings of the IEEE/CVF International Conference on Computer Vision}, pages 1780--1790, 2021.

\bibitem[Kim et~al.(2021)Kim, Son, and Kim]{kim2021vilt}
Wonjae Kim, Bokyung Son, and Ildoo Kim.
\newblock Vilt: Vision-and-language transformer without convolution or region supervision.
\newblock In \emph{International Conference on Machine Learning}, pages 5583--5594. PMLR, 2021.

\bibitem[Lester et~al.(2021)Lester, Al-Rfou, and Constant]{lester2021power}
Brian Lester, Rami Al-Rfou, and Noah Constant.
\newblock The power of scale for parameter-efficient prompt tuning.
\newblock \emph{arXiv preprint arXiv:2104.08691}, 2021.

\bibitem[Li et~al.(2019)Li, Yatskar, Yin, Hsieh, and Chang]{li2019simple}
LH~Li, M~Yatskar, D~Yin, CJ~Hsieh, and KW~Chang.
\newblock A simple and performant baseline for vision and language.
\newblock \emph{arXiv preprint arXiv:1908.03557}, 2019.

\bibitem[Li and Liang(2021)]{li2021prefix}
Xiang~Lisa Li and Percy Liang.
\newblock Prefix-tuning: Optimizing continuous prompts for generation.
\newblock \emph{arXiv preprint arXiv:2101.00190}, 2021.

\bibitem[Li et~al.(2021)Li, Liang, Zhao, Cui, Ouyang, Shao, Yu, and Yan]{li2021supervision}
Yangguang Li, Feng Liang, Lichen Zhao, Yufeng Cui, Wanli Ouyang, Jing Shao, Fengwei Yu, and Junjie Yan.
\newblock Supervision exists everywhere: A data efficient contrastive language-image pre-training paradigm.
\newblock \emph{arXiv preprint arXiv:2110.05208}, 2021.

\bibitem[Radford et~al.(2021)Radford, Kim, Hallacy, Ramesh, Goh, Agarwal, Sastry, Askell, Mishkin, Clark, et~al.]{radford2021learning}
Alec Radford, Jong~Wook Kim, Chris Hallacy, Aditya Ramesh, Gabriel Goh, Sandhini Agarwal, Girish Sastry, Amanda Askell, Pamela Mishkin, Jack Clark, et~al.
\newblock Learning transferable visual models from natural language supervision.
\newblock In \emph{International conference on machine learning}, pages 8748--8763. PMLR, 2021.

\bibitem[Razdaibiedina et~al.(2023)Razdaibiedina, Mao, Hou, Khabsa, Lewis, Ba, and Almahairi]{razdaibiedina2023residual}
Anastasia Razdaibiedina, Yuning Mao, Rui Hou, Madian Khabsa, Mike Lewis, Jimmy Ba, and Amjad Almahairi.
\newblock Residual prompt tuning: Improving prompt tuning with residual reparameterization.
\newblock \emph{arXiv preprint arXiv:2305.03937}, 2023.

\bibitem[Xian et~al.(2017)Xian, Schiele, and Akata]{xian2017zero}
Yongqin Xian, Bernt Schiele, and Zeynep Akata.
\newblock Zero-shot learning-the good, the bad and the ugly.
\newblock In \emph{Proceedings of the IEEE conference on computer vision and pattern recognition}, pages 4582--4591, 2017.

\bibitem[Yao et~al.(2021{\natexlab{a}})Yao, Huang, Hou, Lu, Niu, Xu, Liang, Li, Jiang, and Xu]{yao2021filip}
Lewei Yao, Runhui Huang, Lu~Hou, Guansong Lu, Minzhe Niu, Hang Xu, Xiaodan Liang, Zhenguo Li, Xin Jiang, and Chunjing Xu.
\newblock Filip: Fine-grained interactive language-image pre-training.
\newblock \emph{arXiv preprint arXiv:2111.07783}, 2021{\natexlab{a}}.

\bibitem[Yao et~al.(2021{\natexlab{b}})Yao, Zhang, Zhang, Liu, Chua, and Sun]{yao2021cpt}
Yuan Yao, Ao~Zhang, Zhengyan Zhang, Zhiyuan Liu, Tat-Seng Chua, and Maosong Sun.
\newblock Cpt: Colorful prompt tuning for pre-trained vision-language models.
\newblock \emph{arXiv preprint arXiv:2109.11797}, 2021{\natexlab{b}}.

\bibitem[Zhou et~al.(2022{\natexlab{a}})Zhou, Yang, Loy, and Liu]{zhou2022conditional}
Kaiyang Zhou, Jingkang Yang, Chen~Change Loy, and Ziwei Liu.
\newblock Conditional prompt learning for vision-language models.
\newblock In \emph{Proceedings of the IEEE/CVF Conference on Computer Vision and Pattern Recognition}, pages 16816--16825, 2022{\natexlab{a}}.

\bibitem[Zhou et~al.(2022{\natexlab{b}})Zhou, Yang, Loy, and Liu]{zhou2022learning}
Kaiyang Zhou, Jingkang Yang, Chen~Change Loy, and Ziwei Liu.
\newblock Learning to prompt for vision-language models.
\newblock \emph{International Journal of Computer Vision}, 130\penalty0 (9):\penalty0 2337--2348, 2022{\natexlab{b}}.

\end{thebibliography}
\endgroup

\clearpage
\appendix
\setcounter{table}{0}
\renewcommand{\thetable}{\Alph{section}\arabic{table}}
\setcounter{figure}{0}
\renewcommand{\thefigure}{\Alph{section}\arabic{figure}}

This appendix is organized as follows:
\begin{itemize}
  \item Section A provides the experimental and dataset details for PRE.
  \item Section B provides extended Ablation studies about the effect of the context length, prompt initialization, and parameter sharing on PRE of the base-to-new generalization experiments. It also studies different residual network architectures in PRE.
  \item  Section C provides the interpretation of the learned textual prompts using the nearest words in the embedding space.
  \item Section D gives additional detailed results for each dataset of the base-to-new generalization experiments in different K-shot settings and the detailed experiment results to see the impact of the residual network as well as parameter sharing settings on PRE.
\end{itemize}

\subsection{EXPERIMENTAL DETAILS}\label{AA}
The detailed instructions on how to run the PRE method to reproduce the results are available at the GitHub link: \href{https://github.com/minhanh151/PRE}{github.com/minhanh151/PRE}.
\subsubsection{\textbf{Datasets Details}} 
The datasets utilized in our experiments align with the ones employed in the CoOp \cite{zhou2022learning}. These datasets encompass 8 different benchmarks designed for few-shot visual recognition. For comprehensive reference, Table A1 provides detailed information about each dataset, such as the number of classes, sizes of the training and testing sets, and the original image recognition tasks associated with each dataset.

\subsubsection{\textbf{Training Details:}} 
Adopting the training settings from CoOp \cite{zhou2022learning}, we maintain a consistent training epoch of 50 for all the experiments conducted with various shots. For prompt-based models, we employ a batch size of 32, except in the case of CoCoOp. As reported by (Zhou et al., 2022) \cite{zhou2022conditional}, CoCoOp exhibits a considerable GPU memory consumption when the batch size is set larger than one. Hence, we follow their original configuration and set the batch size to 1 for CoCoOp in our experiments.
We conducted our experiments with two NVIDIA GeForce RTX 3080 Ti GPUs, with 8 GB of memory each. On each task,
training took between 4 minutes and 2 hours.
\subsubsection{\textbf{Hyperparameters}} 
The original CoOp method has different versions with different class token positions and parameter initialization strategies. To maintain consistency, we selected a specific configuration for our baseline, where the token position is "end," the parameter initialization strategy is "a photo of a," and the length of learnable context tokens is 4, similar to the CoOp and CoCoOp settings. 

\begin{table*}[t]
\centering
\caption{Details of 8 Datasets for Few-Shot Visual Recognition and Base-to-New Generalization Image Recognition Evaluation.}
\label{table:A1}
\begin{adjustbox}{width=1\textwidth}
\begin{tabular}{lcccc}
\toprule
\addlinespace[0.95ex]
Dataset & Classes & Training Size & Testing Size & Task \\
\addlinespace[0.5ex]
\midrule
\addlinespace[1.25ex]
Caltech101 (Fei-Fei et al., 2004) & 100 & 4,128 & 2,465 & Object Recognition \\
\addlinespace[0.95ex]
DTD (Cimpoi et al., 2014) & 47 & 2,820 & 1,692 & Texture Recognition \\
\addlinespace[0.95ex]
EuroSAT (Helber et al., 2019) & 10 & 13,500 & 8,100 & Satellite Image Recognition \\
\addlinespace[0.95ex]
FGVCAircraft (Maji et al., 2013) & 100 & 3,334 & 3,333 & Fine-Grained Aircraft Recognition \\
\addlinespace[0.95ex]
Flowers102 (Nilsback \& Zisserman, 2008) & 102 & 4,093 & 2,463 & Fine-Grained Flowers Recognition \\
\addlinespace[0.95ex]
Food101 (Bossard et al., 2014) & 101 & 50,500 & 30,300 & Fine-Grained Food Recognition \\
\addlinespace[0.95ex]
OxfordPets (Parkhi et al., 2012) & 37 & 2,944 & 3,669 & Fine-Grained Pets Recognition \\
\addlinespace[0.95ex]
StanfordCars (Krause et al., 2013) & 196 & 6,509 & 8,041 & Fine-Grained Car Recognition \\
\addlinespace[0.95ex]
\bottomrule
\addlinespace[3.5ex]
\end{tabular}
\end{adjustbox}
\end{table*}

\subsection{EXTENDED ABLATION STUDIES}\label{AA}
\subsubsection{\textbf{Effect of context length}} 
In this study, we examine the significance of the context length for the learnable prompts. To analyze its impact on base-to-new generalization, we conduct experiments using the ViT-16/B backbone with PRE models. Following a similar approach to CoOp \cite{zhou2022learning}, we investigate three context lengths: 4, 8, and 16 context tokens. For context lengths of 8 and 16, the prompt is initialized as "X X ... X a photo of a [Class ]". The performance using PRE across 8 datasets is then averaged and summarized in Fig. A1. Notably, we observe that setting the context length to 8 consistently yields superior performance compared to the other two settings across all three evaluation metrics. Learning prompts with context lengths of 4 and 16 exhibit similar performance levels on PRE. To ensure a fair comparison with CoOp and CoCoOp, we ultimately opt to set the context length to 4 in our final model. This decision ensures consistency in the experimental setup and facilitates a proper evaluation against the previous state-of-the-art methods.

\begin{figure}[t]
    \centering
    \includegraphics[width=0.8\textwidth]{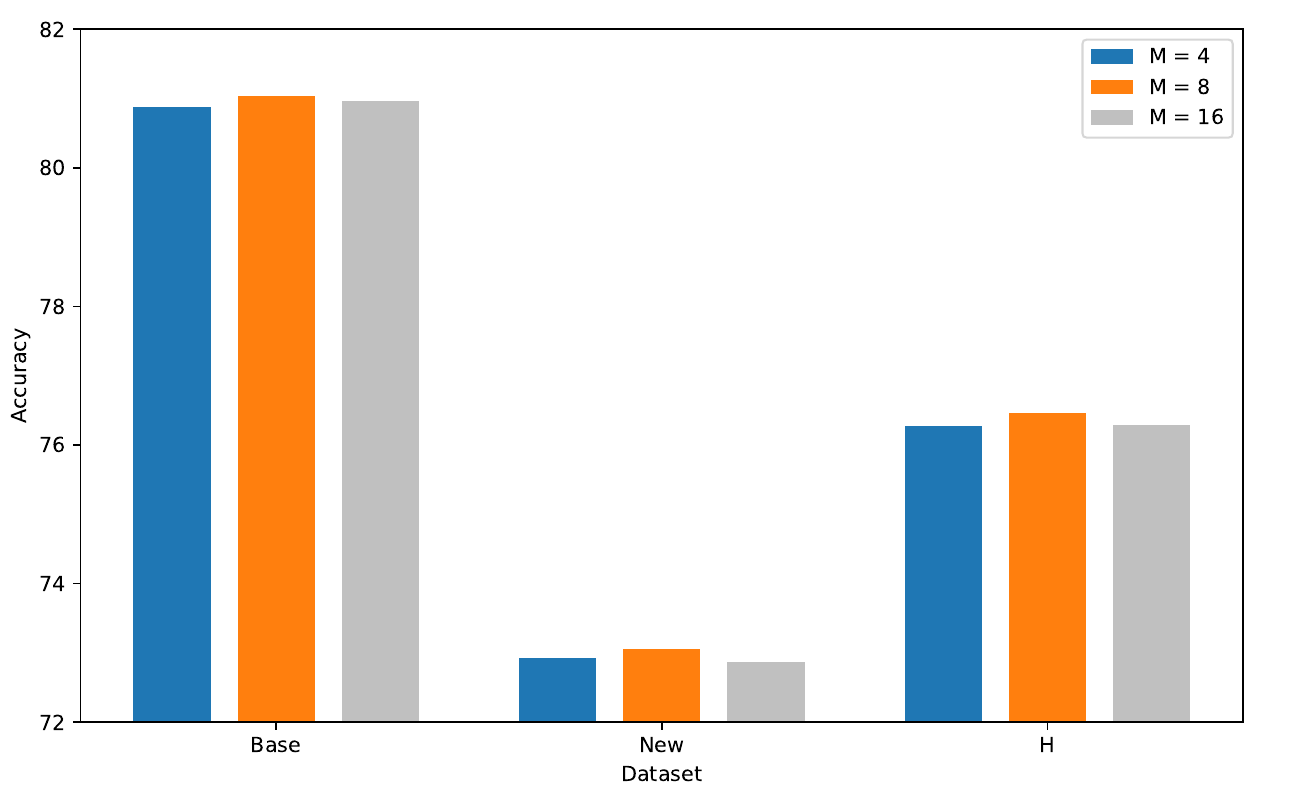}
    \caption{The impact of different context length on average Accuracy on \textit{Base}, \textit{New} and \textit{H} over 8 datasets using PRE model.}
    \label{fig:my_label}
\end{figure}

\begin{figure}[t]
    \centering
    \includegraphics[width=0.8\textwidth]{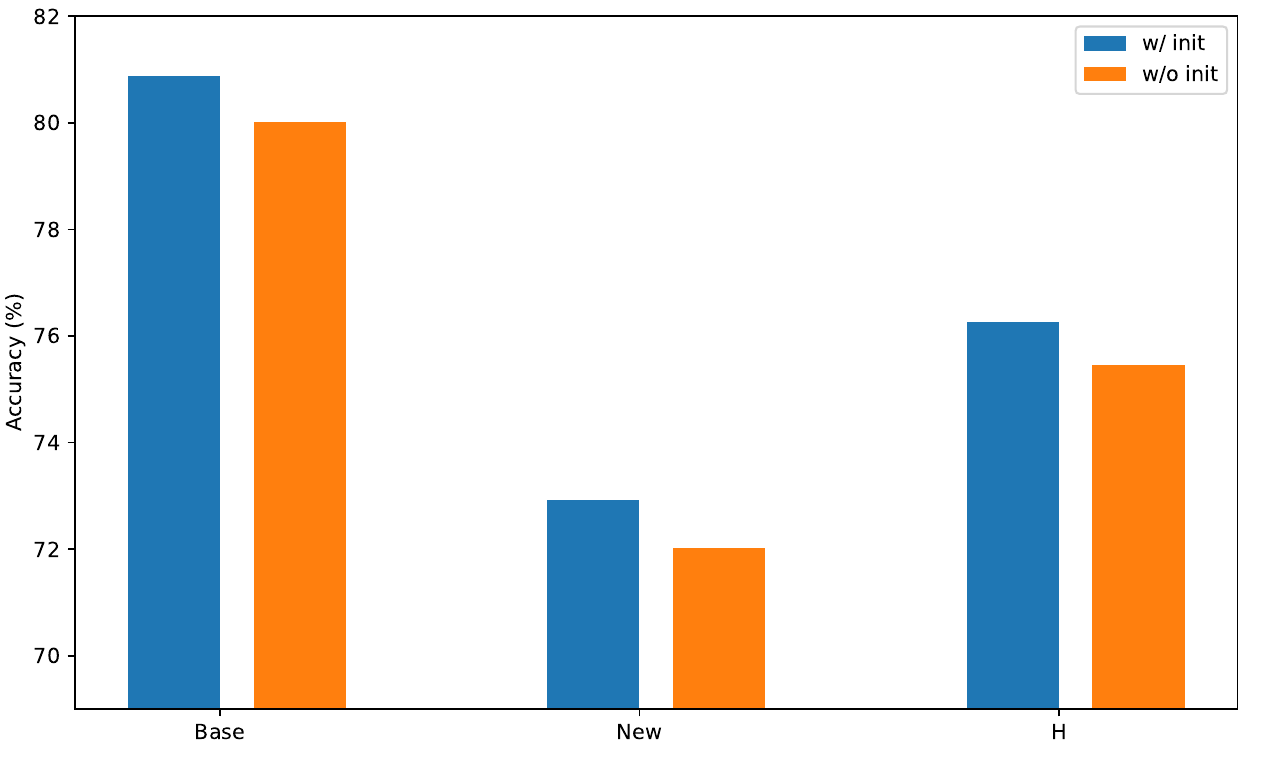}
    \caption{Comparison with vs without prompt initialization on PRE on average Accuracy on \textit{Base}, \textit{New} and \textit{H} over 8 datasets using PRE model.}
    \label{fig:my_label}
\end{figure}

\begin{figure*}[t] 
\centering
\begin{subfigure}{0.40\textwidth}
\centering
\includegraphics[width=\textwidth]{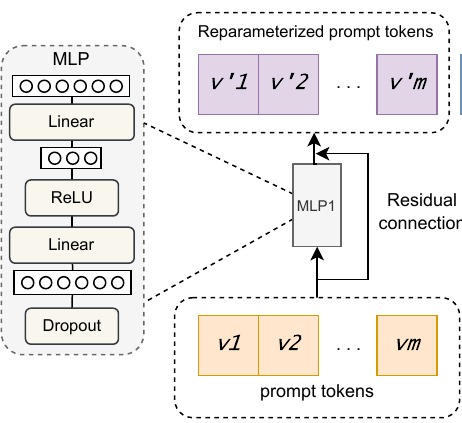}
\caption{A shared bottleneck MLP for every prompt token}
\label{fig:figure1}
\end{subfigure}\hfill
\begin{subfigure}{0.51\textwidth}
\centering
\includegraphics[width=0.76\textwidth]{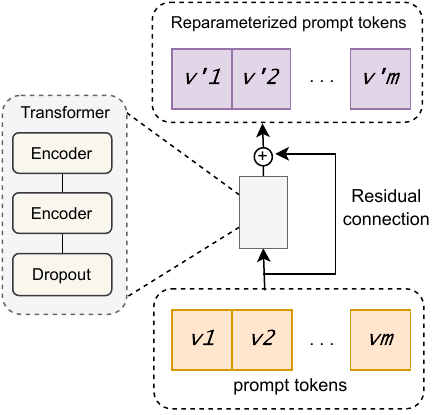}
\caption{A shared Transformer Encoder network for every prompt token}
\label{fig:figure2}
\end{subfigure}
\caption{Two other network architectures for the prompt encoder in PRE.}
\label{fig:both_figures}
\end{figure*}

\subsubsection{\textbf{Effect of prompt initialization}} 
Lester et al. (2021) \cite{lester2021power} find that the initialization of prompt parameters plays a major role in the final
performance. To assess the influence of prompt initialization on prompt tuning, we carry out a comparative analysis employing two different methods: word embeddings-based initialization ('w/ init') and random initialization ('w/o init') on PRE. For random initialization, we utilize a zero-mean Gaussian distribution with a standard deviation of 0.02 to initialize the prompt tokens. On the other hand, word embeddings-based initialization involves initializing the prompt tokens with the phrase "a photo of a." with a context length of 4.
After conducting our experiments and evaluating the performance across 8 datasets, we have summarized the results in Fig. A2. Notably, we have observed that employing word embeddings-based initialization yields slightly higher performance in all three evaluation metrics when compared to random initialization in PRE. This finding highlights the significance of the initialization strategy in prompt tuning and its substantial impact on the overall performance of the model.

\subsubsection{\textbf{Studies of different network architectures for Prompt Encoder}} 
\label{sec:B3}
In this study, we examine other different network architectures for reparameterizing soft prompt embeddings in PRE.
Following Anastasia et al. (2023) \cite{razdaibiedina2023residual}, we use a bottleneck MLP network (as shown in Fig. A3a). Specifically, this MLP network consists of a down-projection linear layer followed by a ReLU layer, and an up-projection linear layer, with a dropout layer at the end. The bottleneck size hyperparameter has been thoroughly studied to get the optimal performance. We also explored a Transformer Encoder network consisting of two Transformer encoder layers followed by a dropout layer, each layer equipped with two attention heads (as in Fig. A3b).
The residual connection is used in all three network architectures (BiLSTM, MLP, and Transformer Encoders network) to combine the input prompt embeddings and the output of the reparameterization network.
The reparameterizing network can be shared among every prompt token or separately process each prompt token using separate parameters. Table A2 shows the results of these networks on the base-to-new generalization setting. These results have been discussed in the Ablation Studies section. Following that, we expand the ablation study to investigate the impact of parameter sharing on these networks in the next part.

\begin{table*}[t]
\centering
\caption{Comparison in PRE using different reparameterizing network architectures in terms of the average \textit{Base}, \textit{New}, and \textit{H} performance in the base-to-new generalization setting.}
\renewcommand{\arraystretch}{1.0}
\begin{tabular}{lcccc}
\toprule
\addlinespace[0.95ex]
\multirow{2}{*}{Encoder Network} & \multirow{2}{*}{\shortstack{Parameter Sharing}} & \multicolumn{3}{c}{Accuracy (\%)} \\
\addlinespace[0.5ex]
\cmidrule(lr){3-5} & & Base & New & H \\
\midrule
\addlinespace[0.95ex]
BiLSTM & \textbf{Share} &82.02 & 71.90 & 76.22 \\
\addlinespace[0.5ex]
{BiLSTM} & Separate & 82.72 & 69.02 & 74.60 \\
\addlinespace[0.5ex]
\midrule
\addlinespace[0.95ex]
{Bottle-neck MLP} & \textbf{Share} & 78.77 & 72.04 & 75.00 \\
\addlinespace[0.5ex]
{Bottle-neck MLP} & Separate & 80.26 & 72.07 & 75.54 \\
\addlinespace[0.5ex]
\midrule
\addlinespace[0.95ex]
Transformer Encoders & Share & 82.05 & 71.55 & 75.66 \\
\addlinespace[0.5ex]
{Transformer Encoders} & \textbf{Separate} & 81.49 & 70.21 & 74.97 \\
\addlinespace[0.5ex]
\bottomrule
\end{tabular}
\end{table*}

\subsubsection{\textbf{Effect of processing prompt tokens independently by separate networks vs. jointly by a shared network in  Prompt Encode}} 
Fig. A3 illustrates the performance comparison of PRE when employing a shared BiLSTM network for every prompt token versus separate BiLSTM networks for each prompt token while Fig. A4 shows the impact of a shared MLP network vs. separate MLP networks on the prompt encoder. The detailed results of parameter sharing on three kinds of network architecture BiLSTM, Transformer Encoders, and MLP are shown in Table 2. We can see that the shared network exhibits significantly better performance on the \textit{Base} classes, while still maintaining strong generalization capabilities. This performance improvement indicates that parameter sharing in the BiLSTM Encoder network of PRE is highly beneficial, as it allows for the effective learning of dependencies between prompt tokens, outperforming the separate network setting.
\begin{figure}[t]
    \centering
    \includegraphics[width=0.8\textwidth]{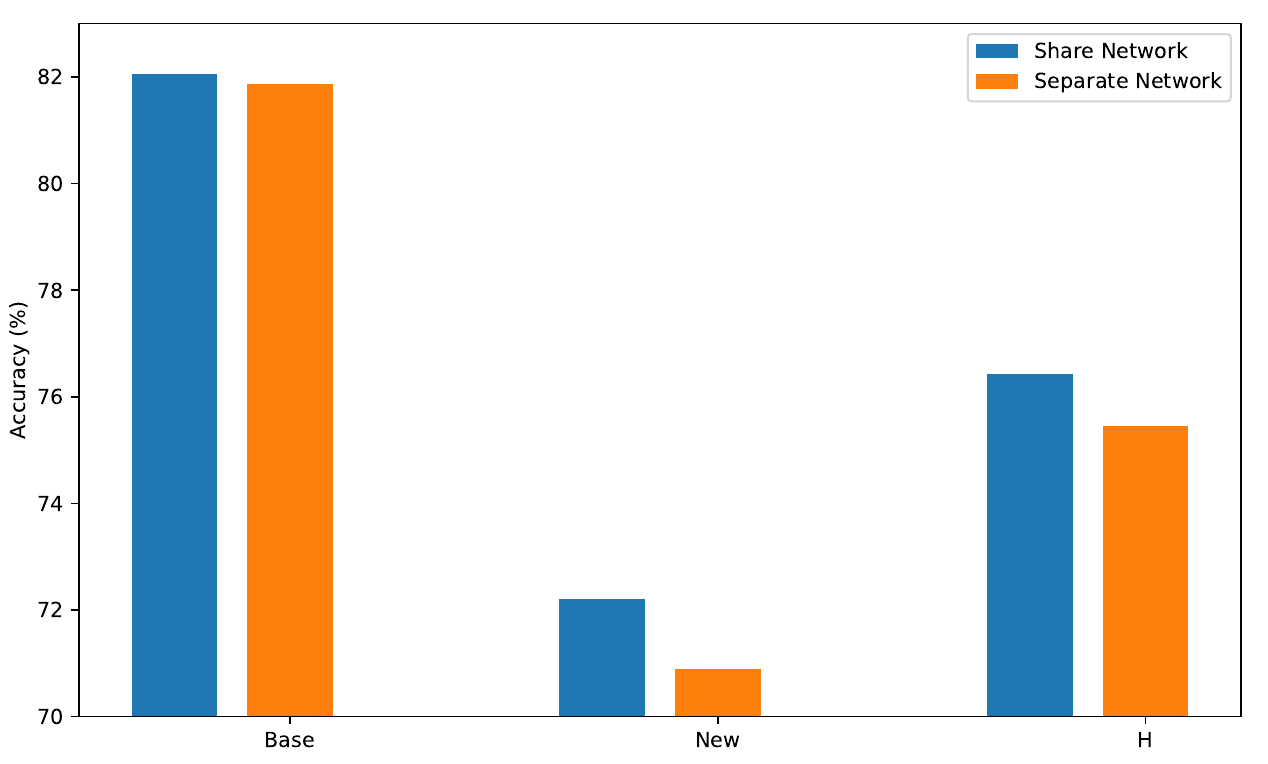}
    \caption{Performance of PRE with shared and separate BiLSTM networks on average Accuracy on \textit{Base}, \textit{New} and \textit{H} over 8 datasets.}
    \label{fig:my_label}
\end{figure}

\begin{figure}[t]
    \centering
    \includegraphics[width=0.8\textwidth]{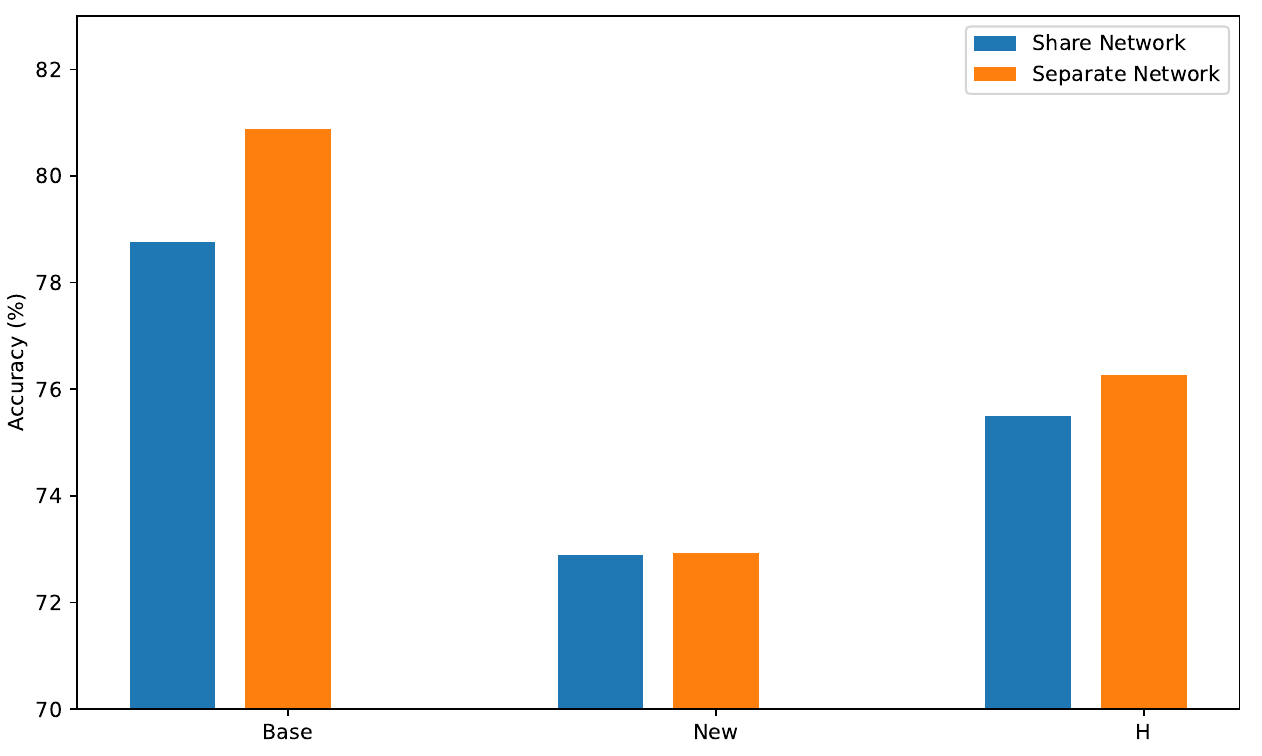}
    \caption{Performance of PRE with shared and separate MLP networks on average Accuracy on \textit{Base}, \textit{New} and \textit{H} over 8 datasets.}
    \label{fig:my_label}
\end{figure}
 Table A3 shows the detailed results on all three network architectures. The downward trend is observed in the case of the Transformer Encoder network as well. This trend is consistent across both network architectures because they both utilize the benefits of processing prompt tokens collectively, which allows them to capture the extended dependencies present within the sequence of prompts.

In contrast, for the MLP network, the separate network configuration yields relatively higher performance on the \textit{Base} classes while maintaining the performance in novel classes. MLP is not affected by the absence of parameter sharing as it does not primarily aim to capture interactions between prompt tokens. In contrast, separate MLP networks with increased parameters can enhance the encoder's adaptability in exploring various mappings of prompt embeddings.

\begin{table*}[h]
\centering
\caption{The nearest words for each of the 4 context vectors learned by PRE with BiLSTM network as prompt encoder, with their distances shown in parentheses.}
\label{tab:scores}
\begin{adjustbox}{width=1\textwidth}
\begin{tabular}{lcccc}
\hline
\addlinespace[1.15ex]
\# & Word 1 & Word 2 & Word 3 & Word 4  \\
\addlinespace[0.75ex]
\hline
\addlinespace[0.95ex]
Caltech101 & borderlands (0.6706) & oth (0.6446) & pose (0.6732) & mark (0.7446) \\
\addlinespace[0.95ex]
OxfordPets & ol (0.5578) & moves (0.6612) & \textbf{wild} (0.6559) & mountain (0.5700)  \\
\addlinespace[0.95ex]
Flowers102 & screen (1.1097) & photoo (1.1657) & battles (1.4090) & sey (1.2732) \\
\addlinespace[0.95ex]
FGVC Aircraft & can (1.4438) & independent (1.8402) & \textbf{tail} (1.7982) & campaigner (1.0018) \\
\addlinespace[0.95ex]
DTD & aster (0.9711) & \textbf{consecutive} (0.8758) & \textbf{line} (1.0512) & stones (1.0743) \\
\addlinespace[0.95ex]
EuroSAT & three (0.4983) & report (0.5567) & rain (0.4577) & pose (0.5220) \\
\addlinespace[0.95ex]
Stanford Cars & salt (0.9016) & riot (1.1503) & N/A (0.9737) & toby (1.1054) \\
\addlinespace[0.95ex]
Food101 & tur (1.1864) & color (0.7646) & lh (1.0958) & water (0.9743) \\
\addlinespace[0.95ex]
\hline
\addlinespace[3ex]
\end{tabular}
\end{adjustbox}
\end{table*}

\begin{table*}[h]
\centering
\caption{The nearest words for each of the 4 context vectors learned by PRE with Transformer Encoder network as prompt encoder. N/A
means non-Latin characters.}
\label{tab:scores}
\begin{adjustbox}{width=1\textwidth}
\begin{tabular}{lcccc}
\hline
\addlinespace[1.15ex]
\# & Word 1 & Word 2 & Word 3 & Word 4  \\
\addlinespace[0.75ex]
\hline
\addlinespace[0.95ex]
Caltech101 & draws (0.5176) & ability (0.4955) & mirrors (0.6053) & pre (0.4966) \\
\addlinespace[0.95ex]
OxfordPets & sep (0.5221) & line (0.6307) & living (0.4938) & marked (0.5395)  \\
\addlinespace[0.95ex]
Flowers102 & bag (0.7056) & paint (0.5929) & pup (0.5806) & pray (0.8519) \\
\addlinespace[0.95ex]
FGVC Aircraft & nail (0.7897) & cranl (1.2032) & mor (0.7339) & wound (0.8062) \\
\addlinespace[0.95ex]
DTD & aster (0.9711) & \textbf{consecutive} (0.8758) & \textbf{line} (1.0512) & stones (1.0743) \\
\addlinespace[0.95ex]
EuroSAT & ma (0.4567) & kt (0.5447) & wong (0.4650) & N/A (0.4693) \\
\addlinespace[0.95ex]
Stanford Cars & mines (0.4830) & line (0.4581) & \textbf{heights} (0.4842) & sheet (0.4345) \\
\addlinespace[0.95ex]
Food101 & buddies (0.4452) & losing (0.7646) & voter (0.5958) & marching (0.4493) \\
\addlinespace[0.95ex]
\hline
\addlinespace[2ex]
\end{tabular}
\end{adjustbox}
\end{table*}

\subsection{INTERPRETATION OF PROMPTS}\label{AA}

The prompts learned through optimization in the continuous space are difficult for humans to understand \cite{zhou2022learning}. To address this issue, CoOp introduces a method that utilizes the nearest words in the embedding space to represent and visualize the learned prompts. In line with this approach, we present the nearest words corresponding to our learned prompts across 8 datasets in Table A3 and Table A4. Like CoOp's findings, most of these words remain difficult to interpret directly through human logic. However, we can still see some connections between these prompts and the corresponding images in certain datasets. For instance, in the OxfordPets dataset, the prompts learned by PRE seem to emphasize elements related to "wild" and "mountain" environments in each image. In the FGVC Aircraft dataset, the learned prompts from PRE show a preference for the tail region to distinguish between different types of airplanes. Moreover, in the DTD dataset, PRE with both types of encoder network focuses on prompts associated with "consecutive" and "line", which appear to be indicative of specific image texture characteristics.  It demonstrates that the learned multiple prompts focus on particular attributes of categories.

\subsection{DETAILED RESULTS}\label{AA}
This section presents a detailed comparison of the proposed
PRE approach with existing methods in
various aspects. Table A5 shows the results in the base-to-new generalization setting of PRE using a shared network and separate network as the reparameterizing mechanism for input prompt embeddings. 

\begin{table*}[htbp]
\centering
\caption{Detailed results on 8 datasets to see the impact of residual connection on different encoder architectures of PRE in the base-to-new generalization setting with ViT-B/16 as the backbone. The
context length M is 4 with the 16-shot samples from the base classes.}
\renewcommand{\arraystretch}{1.2}
\begin{adjustbox}{width=1\textwidth}
    \begin{tabular}{lcc|c|c|c|c|c|c}
\toprule
\addlinespace[1.5ex]
\textbf{Dataset} & \textbf{Set} & \textbf{CoOp} & \multicolumn{2}{c}{\textbf{BiLSTM}} & \multicolumn{2}{c}{\textbf{Transformer Encoders}} & \multicolumn{2}{c}{\textbf{MLP}}\\
\cmidrule(lr){4-9} & & & \textbf{Residual} & W/o Residual & \textbf{Residual} & W/o Residual & \textbf{Residual} & W/o Residual  \\
 \addlinespace[0.5ex]
\midrule
 \addlinespace[0.5ex]
\multirow{3}{*}{Average} & Base & 83.32 & 82.02 & 80.51 & 82.05 & 82.01 & 78.77 & 74.54 \\
 \addlinespace[0.5ex]
 & New & 66.92 & \textbf{71.90} & 70.57 & \textbf{71.55} & 69.34 & \textbf{72.04} & 68.88\\
  \addlinespace[0.5ex]
 & H & 73.34 & \textbf{76.22} & 74.67 & \textbf{75.98} & 74.54 & \textbf{75.00} & 71.16 \\
\midrule
 \addlinespace[0.5ex]
\multirow{3}{*}{Caltech101} & Base & 98.11 & 98.00 & 97.77 & 98.00 & 98.00 & 97.63 & 97.10 \\
 \addlinespace[0.5ex]
 & New & 93.02 & 93.50 & 93.90 & 93.40 & 92.10 & 94.17 & 94.200 \\
  \addlinespace[0.5ex]
 & H & 95.50 & 95.70 & 95.80 & 95.64 & 94.96 & 95.87 & 95.63 \\ \midrule
  \addlinespace[0.5ex]
\multirow{3}{*}{OxfordPets} & Base & 94.24 & 95.27 & 94.90 & 95.37 & 96.50 & 95.37 & 95.30\\
 \addlinespace[0.5ex]
 & New & 96.66 & 97.60 & 97.13 & 97.00 & 95.30 & 97.59 & 97.50 \\
  \addlinespace[0.5ex]
 & H & 95.43 & 96.42 & 96.00 & 96.18 & 95.90 & 96.47 & 96.39 \\ \midrule
  \addlinespace[0.5ex]
\multirow{3}{*}{Flowers102} & Base & 97.63 & 96.40 & 96.73 & 96.00 & 96.62 & 89.23 & 86.60 \\
 \addlinespace[0.5ex]
 & New & 66.55 & 70.80 & 68.43 & 70.90 & 67.93 & 72.53 & 74.60 \\
  \addlinespace[0.5ex]
 & H & 79.15 & 81.64 & 80.16 & 81.56 & 79.77 & 80.02 & 80.15 \\ \midrule
  \addlinespace[0.5ex]
\multirow{3}{*}{\shortstack{FGVC\\Aircraft}} & Base & 39.24 & 35.63 & 34.40 & 36.70 & 37.10 & 33.82 & 23.66 \\
 \addlinespace[0.5ex]
 & New & 23.49 & 32.43 & 33.30 & 31.40 & 30.33 & 32.03 & 11.24 \\
  \addlinespace[0.5ex]
 & H & 29.39 & 34.53 & 33.84 & 33.84 & 33.38 & 32.90 & 15.24\\ \midrule
  \addlinespace[0.5ex]
\multirow{3}{*}{DTD} & Base & 80.17 & 77.80 & 74.70 & 78.60 & 79.20 & 72.37 & 50.87 \\
 \addlinespace[0.5ex]
 & New & 47.54 & 53.93 & 54.50 & 52.75 & 51.30 & 54.60 & 52.10 \\
  \addlinespace[0.5ex]
 & H & 59.69 & 63.70 & 63.02 & 63.13 & 62.27 & 62.24 & 51.48 \\ \midrule
  \addlinespace[0.5ex]
\multirow{3}{*}{EuroSAT} & Base & 91.54 & 86.23 & 85.10 & 87.40 & 87.00 & 81.60 & 82.33 \\
 \addlinespace[0.5ex]
 & New  & 54.44 & 64.47 & 56.50 & 63.73 & 57.10 & 62.00 & 60.20 \\
  \addlinespace[0.5ex]
 & H & 68.28 & 73.78 & 67.91 & 73.71 & 68.95 & 70.46 & 69.55\\ \midrule
  \addlinespace[0.5ex]
 \multirow{3}{*}{\shortstack{Stanford\\Cars}} & Base & 76.20 & 75.83 & 71.47 & 73.34 & 72.74 & 70.20 & 71.50 \\
  \addlinespace[0.5ex]
 & New  & 67.14 & 69.90 & 72.77 & 71.87 & 72.53 & 72.93 & 72.20 \\
  \addlinespace[0.5ex]
 & H & 71.38 & 72.74 & 72.11 & 72.60 & 72.86 & 71.54 & 71.85 \\ \midrule
  \addlinespace[0.5ex]
 \multirow{3}{*}{Food101} & Base & 71.38 & 73.20 & 72.11 & 72.60 & 72.86 & 71.54 & 71.85 \\
  \addlinespace[0.5ex]
 & New & 86.5 & 91.46 & 88.02 & 91.36 & 88.12 & 91.02 & 89.02 \\
  \addlinespace[0.5ex]
 & H & 87.95 & 91.21 & 88.52 & 91.16 & 88.27 & 90.47 & 88.97\\ 
  \addlinespace[0.25ex]
\bottomrule
\end{tabular}
\end{adjustbox}
\end{table*}

\end{document}